\documentclass[a4paper]{article}

\usepackage{INTERSPEECH2021}

\usepackage{times}
\usepackage{latexsym}
\usepackage[T1]{fontenc}
\usepackage[utf8]{inputenc}
\usepackage{microtype}

\usepackage{booktabs} 
\usepackage{multirow}

\usepackage{url,graphicx}
\usepackage{babel}
\usepackage{verbatim}
\usepackage{amsfonts}
\usepackage{caption}
\usepackage{subcaption}
\usepackage{xcolor}

\newcommand{\mask}{\texttt{MASK}}
\newcommand{\keep}{\texttt{KEEP}}
\newcommand{\rand}{\texttt{RAND}}
\newcommand{\ins}{\texttt{INSERT}}
\newcommand{\drop}{\texttt{DROP}}
\newcommand{\s}{\hspace{0.7mm}}
\title{Correcting Automated and Manual Speech Transcription Errors using Warped Language Models}

\name{Mahdi Namazifar, John Malik, Li Erran Li, Gokhan Tur, Dilek Hakkani Tür}
\address{
  Amazon Alexa AI}
\email{mahdinam@amazon.com, jmmlik@amazon.com, lilimam@amazon.com, gokhatur@amazon.com, hakkanit@amazon.com}

\author{Mahdi Namazifar \\
  Amazon Alexa AI \\
  \texttt{mahdinam@amazon.com} \\
  \And
  John Malik \\
  Amazon Alexa AI \\
  \texttt{jmmlik@amazon.com} \\
  \And
  Li Erran Li \\
  Amazon Alexa AI \\
  \texttt{lilimam@amazon.com} \\
  \And
  Gokhan Tur \\
  Amazon Alexa AI \\
  \texttt{gokhatur@amazon.com} \\
  \And
  Dilek Hakkani Tür \\
  Amazon Alexa AI \\
  \texttt{hakkanit@amazon.com} \\
 } 
 
\begin{document}

\maketitle
\begin{abstract}
Masked language models have revolutionized natural language processing systems in the past few years. A recently introduced generalization of masked language models called warped language models are trained to be more robust to the types of errors that appear in automatic or manual transcriptions of spoken language by exposing the language model to the same types of errors during training. In this work we propose a novel approach that takes advantage of the robustness of warped language models to transcription noise for correcting transcriptions of spoken language. We show that our proposed approach is able to achieve up to 10\% reduction in word error rates of both automatic and manual transcriptions of spoken language.

\end{abstract}

\section{Introduction}
Automatic Speech Recognition (ASR) has been an indispensable component of human-machine interaction. Despite tremendous improvements in the quality of ASR systems in recent years, they are still imperfect \cite{8462506, 9054345}. 
On the other hand, training ASR systems requires large amounts of human transcribed spoken language. These human transcriptions are expensive to acquire. At the same time human transcriptions of spoken language are often not fully accurate due to human error. 
As a result, to reduce the number of mistakes in human transcriptions, it is common to have spoken sentences transcribed by multiple different human transcribers, which significantly adds to the cost of human transcription.

\begin{figure*}[t]
  \centering
  \begin{tabular}{c|c}
    \centering
      \begin{subfigure}{0.35\textwidth}
        \includegraphics[width=\textwidth]{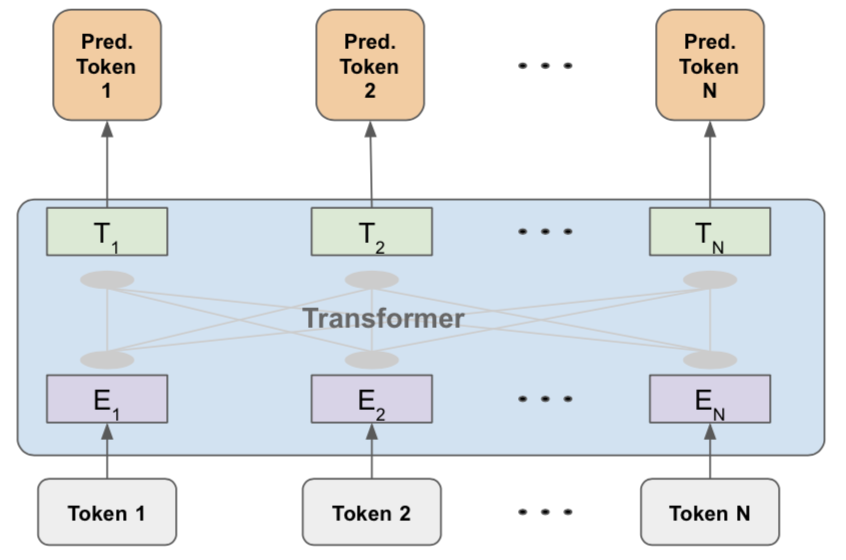}
          \caption{WLM architecture}
          \label{fig:wlm_arch}
      \end{subfigure} &
      \centering
      \begin{subfigure}{0.35\textwidth}
        \includegraphics[width=\textwidth]{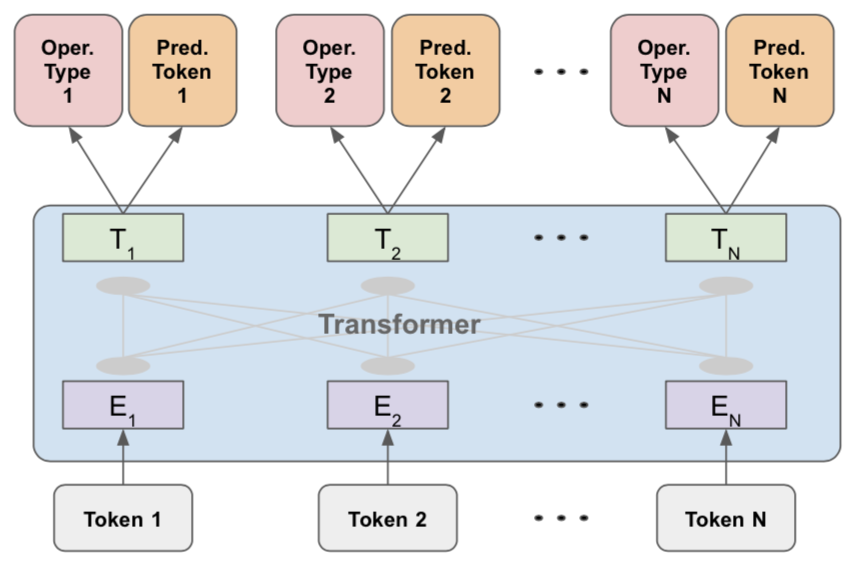}
          \caption{WLM-SC architecture}
          \label{fig:wlm_mod_arch}
      \end{subfigure}
   \end{tabular}
   \caption{\vspace{0mm}The original WLM and WLM-SC model architectures}
   \label{fig:transfer_learning}

\end{figure*}

Warped language models (WLMs) \cite{namazifar2020warped},  variants of masked language models (MLMs) \cite{devlin-etal-2019-bert} are designed to produce encoded representations that are robust to the word-level errors commonly made by ASR systems and human transcribers. That is due to the fact that during the training of warped language models input sentences are corrupted by masking some of the tokens (\texttt{MASK}), replacing some of the tokens with other random tokens (\texttt{RAND}), dropping some of the tokens (\texttt{DROP}), inserting random tokens in random positions in the sentences (\texttt{INSERT}), and keeping some tokens intact (\texttt{KEEP}). Note that these operations (that are referred to as warping operations) resemble the mistakes that are made by ASR systems or by humans during the transcription of spoken language. 
Despite this resemblance, WLMs cannot be directly used for ASR correction due to the lack distinction between different warping operations (more on this in section \ref{sec:mod_wlm}). 
Inspired by the similarity between warping operations of WLMs and mistakes made by ASR systems, and in order to make WLMs work for sentence correction, in this work, we first introduce a modification of warped language models for automatic sentence correction. We refer to this model as WLM sentence correction (WLM-SC). In this model, for each token of the input sentence, a warping operation as well as a token is predicted (Figure \ref{fig:wlm_mod_arch}) from which the corrected sentence could be calculated.  On the other hand, ASR systems often produce additional hypotheses (referred to as ASR n-best list) for the transcription of the spoken language, in addition to the single best hypothesis. These additional hypotheses are rich sources of information on the spoken language.
In this work, we also introduce an approach to incorporate ASR n-best hypotheses in WLM-SC input as additional signals for sentence correction. Our proposed approach for each position in the input sentence predicts what changes should be made, which are independent of the language model of the ASR system. We then use this approach to correct ASR and human transcriptions of spoken language. We show through computational results that our correction approach reduces Word Error Rate (WER) of ASR and human transcriptions of spoken language by up to more than 10\%.

\begin{figure*}[t]
  \centering
  \fbox{\includegraphics[width=0.8\textwidth]{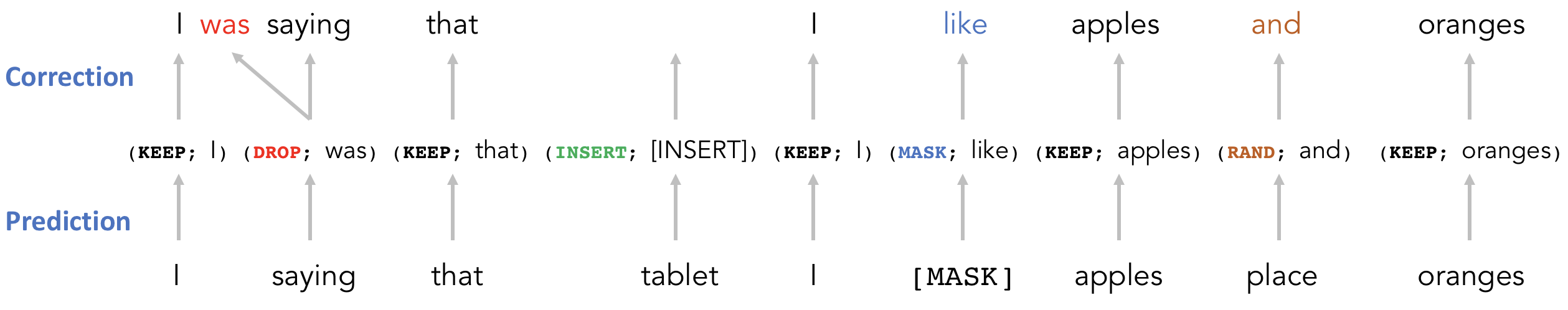}}
          \caption{Example of sentence correction using modified WLM}
          \label{fig:wlm_sc_arch}
  \label{fig:wlm_example}
\end{figure*}

\section{Related Work}

We briefly review related works in correcting ASR and human transcription of spoken language. Human transcription error correction makes use of either the acoustic model or the language model of an ASR system. Acoustic model based methods use the mismatch between the forced alignment of human transcription and posteriors of the acoustic classifier to detect the phonetic errors~\cite{yang-etal-trans-icassp2019, wang-etal-manual2019}.
There have also been several approaches to improve ASR output including hypothesis re-scoring, post-processing and leveraging high level semantic information with joint ASR and Spoken Language Understanding (SLU) models~\cite{weng-etal-icassp2020, Ponnusamy-etal-aaai2020, zheng-etal-iscslp2016, chen-etal-icassp2020, song-etal-2019L2RS}. 
In \cite{Jung2004SpeechRE} the authors propose an ASR correction approach based on maximum entropy language models to correct semantic and lexical errors. In \cite{MANGU2000373,1198851} ASR lattices are converted to word confusion networks, and word error rates are improved by finding consensus across alternative paths in the lattice. More recent approaches such as \cite{8461974} use neural network based language models for ASR correction. The language model based sentence correction methods detect errors by finding the difference between ASR hypotheses and transcription~\cite{hazen-interspeech2006, wang-etal-manual2019}.  
In contrast, in this work we introduce a modification of warped language models trained with high quality ground truth transcription that for each position in the input sentence predicts what changes should be made, which are independent of the language model of the ASR system.

\vspace{-3mm}
\section{Warped LM Sentence Correction}
\label{sec:mod_wlm}

Masked language models are trained by introducing certain types of noise to the input while the model is asked to recover the original input. More specifically, in training masked language models, first a subset of tokens of the input sentences are randomly selected. Some of these tokens  go through the \mask \s operation, some go through the \rand \s operation, and some are left intact (\keep \s operation) \cite{devlin-etal-2019-bert}. The model is then trained to predict the original token for each of the tokens in this subset, i.e. for a masked token, it should predict what was the token that was masked, for a token that was replaced by another random token it should predict the original token, and for a token that was kept intact it should predict the token itself.

\begin{figure}[t!]
    \centering
    \fbox{\includegraphics[width=0.41\textwidth]{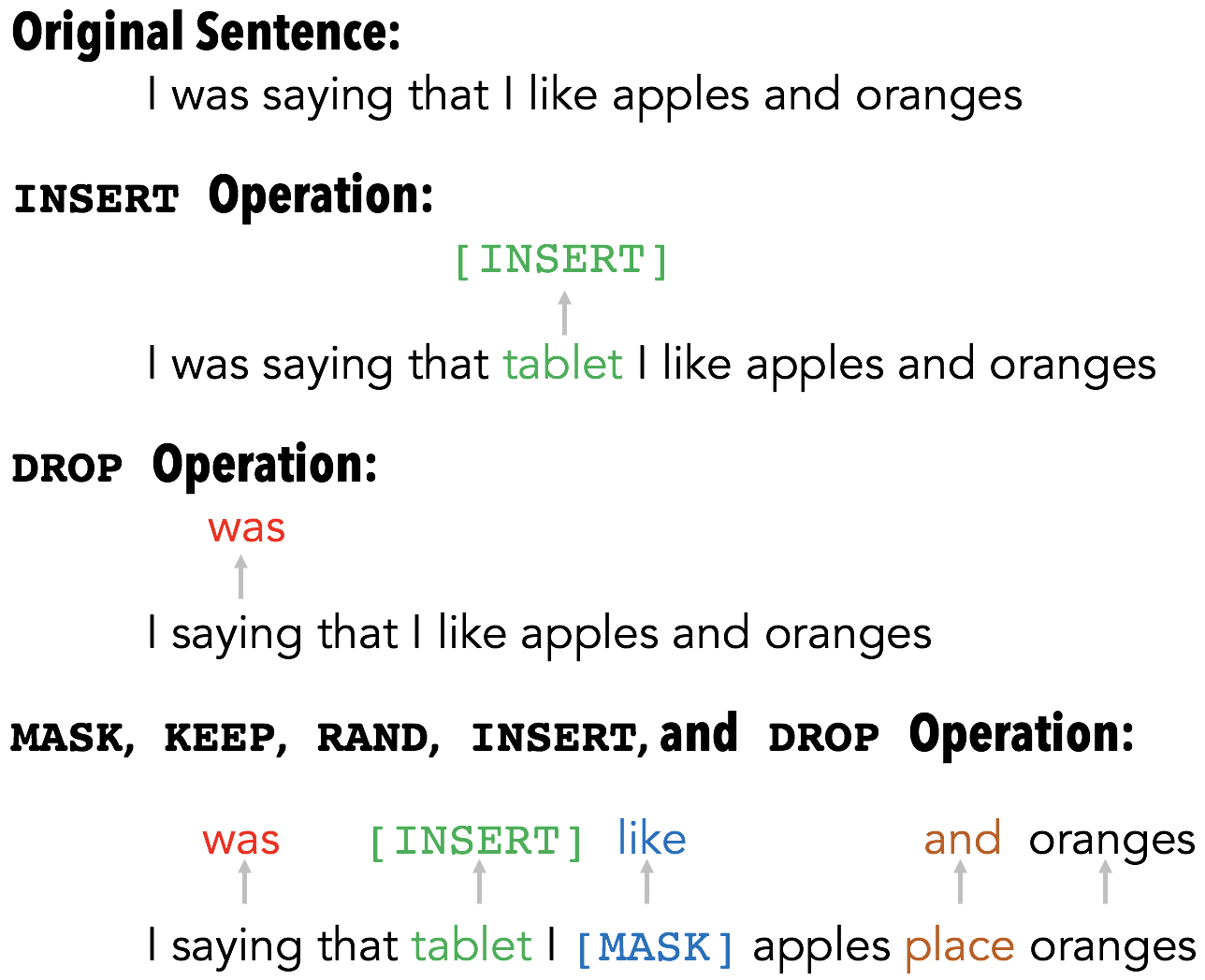}}
    \caption{Example of warping in warped language models borrowed from \cite{namazifar2020warped}.}
    \label{fig:warp}
    \vspace{-4mm}
\end{figure}

Warped language models \cite{namazifar2020warped} are identical to masked language models except that they use two additional operations during training to introduce noise to the input sentences. The first operation is \drop \s for which from the subset of the selected tokens some of the tokens are dropped and the model needs to predict the dropped tokens as the target for the token that comes after the dropped token. The second new operation used in warped language models is \ins \s for which some random tokens are inserted in the input sequence of tokens and for the inserted tokens the model needs to predict the special token \texttt{[INSERT]} as the target. Applying the set of \mask \s, \drop \s, \ins \s, \rand, and \keep \s (referred to as warping operations) to a sentence constitutes what is referred to as warping a sentence. Figure \ref{fig:warp} shows an example of such warping of a sentence.

Considering these warping operations, 
it is easy to show that any ASR or human transcription that includes mistakes could be achieved by applying a limited number of warping operations on the correct transcription. As a result using warped language models for sentence correction intuitively seems like a viable option. One hurdle to do so however is the fact that although warped language models can predict a token for every position in the input sentence, they cannot fully distinguish between all the different warping operations. For instance if for a position the model predicts a token that is different from the original token at that position, it would not be clear whether the model believes that there was a \rand \s operation at that position or a \drop \s operation at the previous position. In order to have a sentence correction model it is required to have not only the tokens for each position, but also the warping operation at every position. Hence we introduced a modified version of warped language models that does exactly that. Figure \ref{fig:wlm_mod_arch} depicts the architecture of this model that we call \textit{WLM Sentence Correction (WLM-SC)}.

\begin{figure*}[h]
  \centering
  \includegraphics[width=0.8\textwidth]{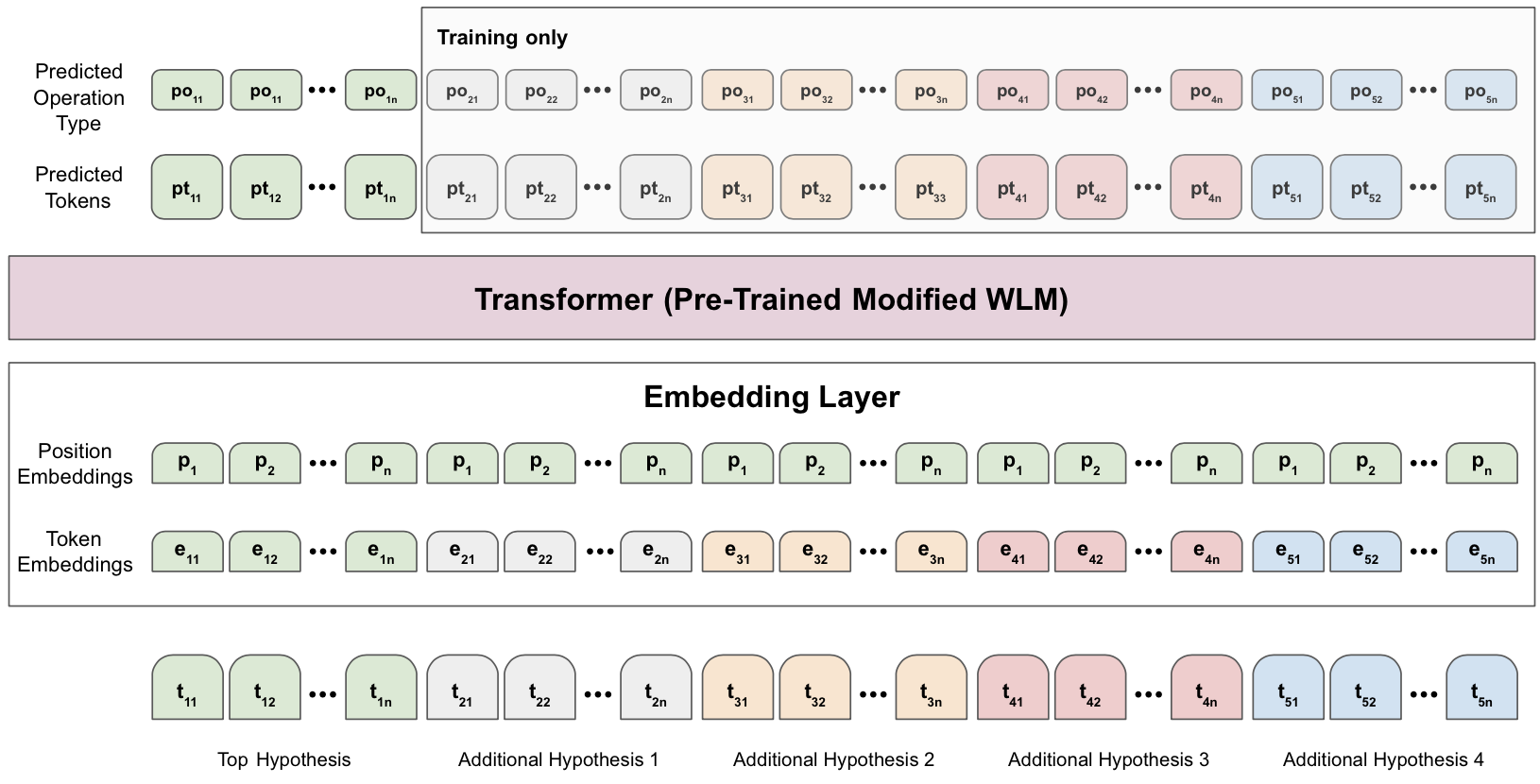}
          \caption{WLM-SC with additional hypotheses}
  \label{fig:wlm_corr_asr}
  \vspace{-4mm}
\end{figure*}

The warping operation prediction head in WLM-SC is a 5-way classifier that predicts one of the warping operations for each position that we predict a token for. To train this model, similar to the original warped language models, first input sentences are warped, i.e. warping operations are applied to a subset of positions in the input sentences. Then the model is trained such that for each of these position it should predict a token and a warping operation that was applied at that position. The training loss is the sum of the token prediction loss and the warping operation prediction loss. At evaluation time, the model predicts a token and a warping operation for every position of input sentences from which corrected sentences could be obtained. Figure \ref{fig:wlm_example} shows an example of how the proposed sentence correction algorithm works.

\section{WLM-SC with ASR Hypotheses}
\label{sec:wlm_with_asr}

\begin{table*}[h]
    \centering
    \begin{tabular}{|c|r||r|r|r||r|r|r|}
    \hline
        & &\multicolumn{3}{c||}{ASR Transcription WER} & \multicolumn{3}{c|}{Human Transcription WER}\\
        \hline
        ASR Confidence & Bin Size & Original & Corrected & Rel. Diff.& Original & Corrected & Rel. Diff.\\
        Score & & & & \% & & & \%\\
        \hline
        $[0.0, 0.2]$ & 12074 & 38.17 & \textbf{37.26} & 2.4& 29.01 & \textbf{28.28} & 2.5 \\
        $(0.2, 0.4]$ & 12280 & 23.99 & \textbf{22.49} & 6.3 & 19.11 & \textbf{17.91} & 6.3\\
        $(0.4, 0.6]$ & 14208 & 16.26 & \textbf{14.66} & 9.8 & 14.03 & \textbf{12.52} &10.8\\
        $(0.6, 0.8]$ & 24463 & 10.64 & \textbf{9.40} & 11.7 & 10.15 & \textbf{8.31} &18.1\\
        $(0.8, 1.0]$ & 59050 & 5.19  & \textbf{4.95}  & 4.6 & 5.56  & \textbf{4.34} &21.9\\
        \hline
    \end{tabular}
    \caption{Stratification of the test set by ASR model confidence score}
    \label{tbl:Results_2}
    \vspace{-5mm}
\end{table*}

In the previous section, we outlined WLM-SC which is based on a modification of warped language models. In this framework, the model receives a sentence as input and outputs a corrected sentence. One use case of sentence correction is when we deal with transcriptions of spoken language. These transcriptions could come from an ASR system or could be done by humans. In both cases these transcriptions are noisy (i.e. contain mistakes) which makes the case for using sentence correction algorithms. For this purpose these transcriptions could directly be fed into the WLM-SC. However, in the case of ASR transcriptions, often times ASR output also includes ASR n-best hypotheses, which could be seen as additional information that could be used for correcting the ASR transcription. On the other hand, for the case of human transcriptions, since running ASR on spoken language is relatively low cost, for correcting human transcriptions of spoken language we also get the ASR hypotheses of the spoken language to use them in the correction process. In the rest of this section, we discuss our proposed approach for correcting a noisy transcription of spoken language, which could be ASR top hypothesis or human transcription (referred to both as \textit{top hypothesis}), with additional ASR hypotheses (referred to as \textit{additional hypotheses})\footnote{In case of human annotation these are ASR top hypothesis along with additional ASR hypotheses}.

The WLM-SC model that was introduced in the previous section (Figure \ref{fig:wlm_mod_arch}) is fine-tuned to perform error corrections for the \textit{top hypothesis} with the additional information that is provided by the \textit{additional hypotheses} (we use up to 4 \textit{additional hypotheses}). Each hypothesis is considered as a warping of the ground truth of the spoken utterance. Additional hypotheses are first aligned with the top hypothesis. 
Then for each hypothesis the minimum number of \ins, \drop, and \rand \s warping operations needed to transform the hypothesis to the ground truth transcription is calculated and are used as labels for fine tuning of the WLM-SC model. Note that instead of randomly creating warping operations here we use the warping operations that exist in the noisy input. All the hypotheses (top and additional) and their labels are first padded to a uniform length and then are concatenated. The input then is embedded using token and position embedding \cite{NIPS2017_3f5ee243} as usual. The only difference here is that the  position embedding layer of the  pre-trained WLM-SC model is  modified  so that the \textit{i}-th token of each tokenized hypothesis is given the same position vector for transformers position embeddings, as the \textit{i}-th token of the tokenized transcription (note that in the ``Position Embeddings'' layer in Figure \ref{fig:wlm_corr_asr} the same sequence of vectors is repeated). The position vectors are then  allowed to be learned independently during training. At training time the loss is calculated over all tokens (for all of the hypotheses). During inference however the prediction is done only for the top hypothesis. Figure \ref{fig:wlm_corr_asr} depicts how ASR hypotheses are incorporated into the WLM-SC model.

\section{Computational Results}

As was mentioned earlier, human transcriptions of spoken language are noisy (include mistakes). In order to establish a ground truth for our experiments, we use  the consensus human transcription of at least three independent transcribers. We refer to these consensus transcriptions as \textit{golden transcriptions} and we assume that golden transcriptions are error free. To evaluate the effectiveness of WLM-SC with additional hypotheses for ASR outputs and human transcriptions of spoken language, we calculate the word error rate (WER) metric over a held out test set of 100,000  utterances that have golden transcriptions. Specifically, to calculate WER the sum of the Levenshtein distances between all of the candidate corrections and their corresponding golden transcriptions is divided by the sum of the lengths of all of the golden transcriptions. The ASR system used in this work is one of our internal ASR systems.

\begin{table*}[h]
    \centering
    \begin{tabular}{|l|r|r|r|r|r|}
    \hline
    & & \multicolumn{4}{c|}{Transcription WER} \\
    \hline
    ASR WER Bin & Bin Size & ASR & ASR WLM-SC & Human & Human WLM-SC \\
    & & & + Add. Hyp. & & + Add. Hyp.\\
    \hline
    \hline
    0.00 & 81185 & 0.00 & 0.35 & 4.65 & 2.36 \\
    \hline
    (0.00, 0.25) & 16151 & 16.80 & 14.87 & 12.83 & 12.04 \\
    \hline
    [0.25, 0.50) & 13205 & 38.87 & 34.98 & 25.08 & 25.37 \\
    \hline
    [0.50, 1.00) & 9764 & 80.56 & 78.32 & 45.59 & 47.03 \\
    \hline
    [1.00, $\infty$] & 2963 & 238.46 & 228.75 & 132.36 & 145.22 \\
    \hline
    \end{tabular}
    \caption{Word error rate (WER) percentage for ASR and human transcription, as well as for WLM-SC with additional hypotheses sentence correction broken down for ASR word error rates. The numbers show that improvements in ASR transcription WER are approximately uniform across ASR WER bins, while improvements in human transcription WER are observed primarily when the ASR hypothesis is correct and can be leveraged to correct word errors.}
    \label{tbl:results_3}
    \vspace{-7mm}
\end{table*}

We first train the modified WLM (WLM-SC) that was introduced in Section \ref{sec:mod_wlm} on the English Wikipedia corpus. Next we fine tune this model with the training data of spoken language transcriptions (ASR or human transcriptions) in the architecture that we introduced in Section \ref{sec:wlm_with_asr} (Figure \ref{fig:wlm_corr_asr}). We calculate the WER of the corrected test set sentences both with WLM-SC as well as with WLM-SC with additional hypotheses, and compare them with the original WER of transcriptions (ASR and human) before the correction. The numbers are shown in Table \ref{tbl:results_1}. It should be mentioned that oracle WER of the ASR n-best (i.e. if the hypothesis from the ASR n-best with the lowest WER is picked) is 11.17. From the table we can see that for ASR transcriptions the WER is reduced from 15.11\% to 14.32\% for WLM-SC, and to 14.24\% for WLM-SC with additional hypotheses (a relative reduction of 4.6\%).  On the other hand, applying WLM-SC on human transcriptions in fact slightly increases WER from 12.38\% to 12.44\%. However when we apply WLM-SC with additional hypotheses on human transcriptions WER is reduced from 12.38 to 11.07, which is a relative improvement of around 10.6\%. Another way to interpret these numbers is that when humans make mistakes, the transcriptions are still fine according to the WLM-SC model. However, these transcriptions may not be fine according to the acoustics information that is captured in the additional ASR hypotheses (ASR n-best).

\begin{table}[t]
    \centering
    \begin{tabular}{|l||c|c|c|}
    \hline
    & Original & WLM-SC & WLM-SC \\
    & & & + Add. Hyp. \\
    \hline
    \hline
    ASR & 15.11 & 14.32 & \textbf{14.24}\\
    \hline
    Human & 12.38 & 12.44 & \textbf{11.07}\\
    \hline
    \end{tabular}
    \caption{Word error rate (WER) percentage between candidate transcriptions in the test set and their golden transcriptions. Oracle WER of the ASR n-best (which means if the hypothesis from the ASR n-best with the lowest WER is picked) is 11.17.}
    \label{tbl:results_1}
    \vspace{-7mm}
\end{table}

In Table~\ref{tbl:Results_2}, we show WER of  the WLM-SC with additional hypotheses applied to ASR and human transcriptions partitioned according to the ASR model confidence score of the ASR 1-best hypothesis. These confidence scores (values between 0 and 1) are partitioned into 5 bins of same length. We observe WER improvement across all groups of utterances. The average relative improvement is $6.95\%$ for ASR 1-best correction and $11.93\%$ for human transcription correction. It is notable that for ASR confidence scores between 0.6 and 0.8 our approach reduces WER for ASR transcriptions by 11.7\% which is highest among all bins. For human transcription, the highest improvement is when the ASR confidence scores are between 0.8 and 1.0 where the overall WER drops from 5.56 to 4.34, which is 21.9\% improvement.


In another set of results we study improvements made by WLM-SC with additional hypotheses for utterances with different ASR WER ranges. Table \ref{tbl:results_3} shows WER percentages for ASR and human transcriptions as well as WER percentages for ASR and human transcriptions corrections with WLM-SC with additional hypotheses. The numbers in this table show that improvements in ASR transcription WER are observed across all ASR WER bins, while improvements in human transcription WER are observed primarily when the ASR hypothesis is correct and can be leveraged to correct word errors. The numbers of the last row of Table \ref{tbl:Results_2} suggest that improvements to human transcriptions could be attributed to ASR additional hypotheses. The results of Table \ref{tbl:results_3} on the other hand suggest that human transcriptions could be improved if the WER of ASR transcriptions are low, i.e. ASR transcriptions are strong hypotheses.

\section{Conclusions}
Transcription of spoken language using Automatic Speech Recognition (ASR) plays a fundamental role in human-machine interaction through natural language. Errors in ASR may result in further issues in natural language understanding and degrade the user-perceived quality of interactions. These ASR systems themselves are trained using manual transcriptions of spoken language done by humans, which often include mistakes due to human errors. Therefore, improving the quality of human transcriptions of spoken language could improve the quality of ASR systems, which in turn positively impacts users' experience when interacting with machines.

In this work we establish a strong connection between the operations that are used to introduce noise in the training of WLMs and the mistakes that are made in transcriptions of spoken language done by humans or by ASR systems. Based upon this connection we introduce WLM-SC as a modification of WLM that are language models that could be directly used for sentence correction. We then introduce a novel approach to incorporate ASR hypotheses for correcting transcriptions (both human and ASR) of spoken language using WLM-SC. In our experimental results we show that our proposed approach not only is capable of improving word error rate of ASR systems, it even reduces word error rate of human transcriptions by up to 10\%. Moreover our results suggest that the gains on human transcription corrections generally come from the additional ASR hypotheses (which could be interpreted as signals from the audio) and are only realized when ASR hypotheses are of higher quality.

\bibliography{main}
\bibliographystyle{IEEEtran}

\end{document}